\documentclass{article}
\usepackage{ijcai18}

\usepackage{times}
\usepackage{xcolor}
\usepackage{soul}
\usepackage[utf8]{inputenc}
\usepackage[small]{caption}

\usepackage{latexsym}
\usepackage{amsmath}
\usepackage{multirow}
\usepackage{url}

\usepackage{graphicx}
\usepackage{CJK}
\usepackage{enumerate}
\usepackage{amssymb}
\usepackage{amsfonts}
\usepackage[noend]{algpseudocode}
\usepackage{algorithmicx,algorithm}
\usepackage{booktabs}

\title{Phrase Table as Recommendation Memory for Neural Machine Translation}

\author{
Yang Zhao $^{1,2}$, 
Yining Wang $^{1,2}$, 
Jiajun Zhang $^{1,2,4}$ and
Chengqing Zong $^{1,2,3}$
\\ 
$^1$ National Laboratory of Pattern Recognition, Institute of Automation, CAS, Beijing, China\\
$^2$ University of Chinese Academy of Sciences, Beijing, China\\
$^3$ CAS Center for Excellence in Brain Science and Intelligence Technology, Beijing, China\\
$^4$ Beijing Advanced Innovation Center for Language Resources, Beijing, China\\
\{yang.zhao, yining.wang, jjzhang, cqzong\}@nlpr.ia.ac.cn
}
\begin{document}
\begin{CJK*}{UTF8}{gbsn}

%\author{Yang Zhao, Yining Wang, Jiajun Zhang and Chengqing Zong\\ 
%National Laboratory of Pattern Recognition, Institute of Automation, CAS, Beijing, China\\
%	     University of Chinese Academy of Sciences, Beijing, China\\
%	     CAS Center for Excellence in Brain Science and Intelligence Technology, Beijing, China\\
%         \{yang.zhao, yining.wang, jjzhang, cqzong\}@nlpr.ia.ac.cn\\
%}

\maketitle

\begin{abstract}
	
Neural Machine Translation (NMT) has drawn much attention due to its promising translation performance recently. However, several studies indicate that NMT often generates fluent but unfaithful translations. In this paper, we propose a method to alleviate this problem by using a phrase table as recommendation memory. The main idea is to add bonus to words worthy of recommendation, so that NMT can make correct predictions. Specifically, we first derive a prefix tree to accommodate all the candidate target phrases by searching the phrase translation table according to the source sentence. Then, we construct a recommendation word set by matching between candidate target phrases and previously translated target words by NMT. After that, we determine the specific bonus value for each recommendable word by using the attention vector and phrase translation probability. Finally, we  integrate this bonus value into NMT to improve the translation results. The extensive experiments demonstrate that the proposed methods obtain remarkable improvements over the strong attention-based NMT.

\end{abstract}

\section{Introduction}

The past several years have witnessed a significant progress in Neural Machine Translation (NMT). Most NMT methods are based on the encoder-decoder
architecture \cite{kalchbrenner2013recurrent,sutskever2014sequence,bahdanau2014neural} and can
achieve promising translation performance in a variety of language pairs \cite{facebook2017convolution,google2017attention,junczys2016neural}.

\begin{figure}[!t]
	\centering
	\includegraphics[width=0.9\columnwidth]{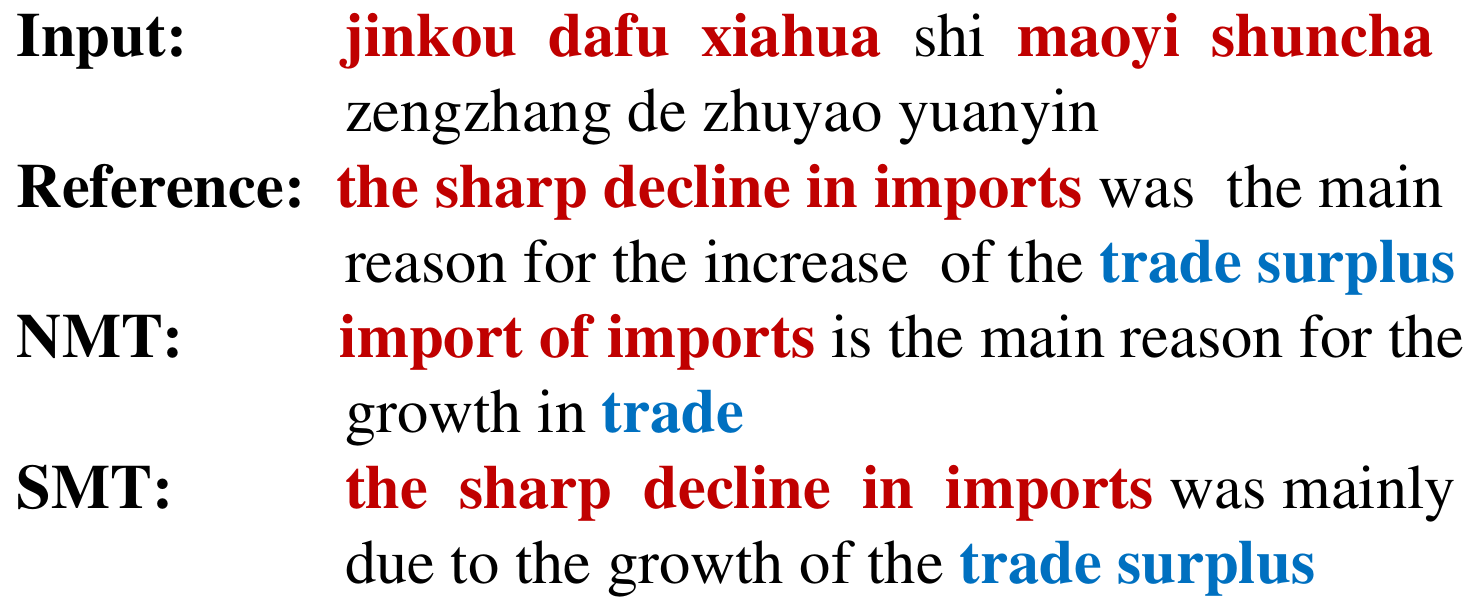}
	\caption{An example of mistakes made by NMT, while SMT can produce a correct translation.}
	\label{overview}
\end{figure}

However, recent studies \cite{arthur2016incorporating,Tu2016Reconstruction} show that NMT often generates words that make target sentences fluent, but unfaithful to the source sentences. In contrast, traditional Statistical Machine Translation (SMT) methods tend to rarely make this kind of mistakes. Fig. 1 shows an example that NMT makes mistakes when translating the phrase ``jinkou dafu xiahua (the sharp decline in imports)'' and the phrase ``maoyi shuncha (the trade surplus)'', but SMT can produce correct results when translating these two phrases.
\cite{arthur2016incorporating} argues that the reason behind this is the use of distributed representations of words in NMT makes systems often generate words that seem natural in the context, but do not reflect the content of the source sentence. Traditional SMT can avoid this problem as it produces the translations based on phrase mappings.

Therefore, it will be beneficial to combine SMT and NMT to alleviate the previously mentioned problem. Actually, researchers have made some effective attempts to achieve this goal. Earlier studies were based on the SMT framework, and have been deeply discussed in \cite{zhang2015deep}. Later, the researchers transfers to NMT framework. Specifically, coverage mechanism \cite{tu2016coverage,mi2016coverage}, SMT features \cite{wang2016neural,he2016improved,stahlberg2016syntactically,li2016towards,Wang2017Towards} and translation lexicons \cite{arthur2016incorporating,zhang2016bridging,feng2017memory} have been fully explored. In contrast, phrase translation table, as the core of SMT, has not been fully studied. Recently, \cite{tang2016neural} and \cite{Wang2017phrases} explore the possibility of translating phrases in NMT. However, the ``phrase'' in their approaches are different from that used in phrase-based SMT. In \cite{tang2016neural}'s models, the phrase pair must be a one-to-one mapping with a source phrase having a unique target phrase (named entity translation pairs). In \cite{Wang2017phrases}'s models, the source side of a phrase pair must be a chunk. Therefore, it is still a big challenge to incorporate any phrase pair in the phrase table into NMT system to alleviate the unfaithfulness problem.

In this paper, we propose an effective method to incorporate a phrase table as recommendation memory into the NMT system. To achieve this, we add bonuses to the words in recommendation set to help NMT make better predictions. Generally, our method contains three steps. 1) In order to find out which words are worthy to recommend, we first derive a candidate target phrase set by searching the phrase table according to the input sentence. After that, we construct a recommendation word set at each decoding step by matching between candidate target phrases and previously translated target words by NMT. 2) We then determine the specific bonus value for each recommendable word by using the attention vector produced by NMT and phrase translation probability extracted from phrase table. 3) Finally we integrate the word bonus value into the NMT system to improve the final results.

In this paper, we make the following contributions:

1) We propose a method to incorporate the phrase table as recommendation memory into NMT system. We design a novel approach to find from the phrase table the target words worthy of recommendation, calculate their recommendation scores and use them to promote NMT to make better predictions.

2) Our empirical experiments on Chinese-English translation and English-Japanese translation tasks show the efficacy of our methods. For Chinese-English translation, we can obtain an average improvement of 2.23 BLEU points. For English-Japanese translation, the improvement can reach 1.96 BLEU points. We further find that the phrase table is much more beneficial than bilingual lexicons to NMT.

\section{Neural Machine Translation}

NMT contains two parts, encoder and decoder, where encoder transforms the source sentence $X=\left \{ x_{1},x_{2},...,x_{Tx}\right \}$ into context vectors $C=\left \{ h_{1},h_{2},...,h_{Tx}\right \}$. This context set is constructed by $m$ stacked Long Short Term Memory (LSTM) \cite{hochreiter1997long} layers. $h_{j}^{k}$ can be calculated as follows:
\begin{equation}
h_{j}^{k}=LSTM(h_{j-1}^{k},h_{j}^{k-1})
\end{equation}

The decoder generates one target word at a time by computing the probability of $p(y_{i}|y_{<i},C)$ as follows:
\begin{equation}
\begin{aligned}
p(y_{i}|y_{<i},C)&=p(y_{i}|y_{<i},c_{i})\\&= softmax(score(W_{y_{i}},\widetilde{z_{i}}))
\end{aligned}
\end{equation}
where $score(W_{y_{i}},\widetilde{z_{i}}))$ is the score produced by NMT:
\begin{equation}
score(W_{y_{i}},\widetilde{z_{i}})=W_{y_{i}}\widetilde{z_{i}}+b_{s}
\end{equation}
and $\widetilde{z_{i}}$ is the attention output:
\begin{equation}
\widetilde{z_{i}}=tanh(W_{c}[z_{i}^{m};c_{i}])
\end{equation}
the attention model calculates $ c_{i}$ as the weighted sum of the source-side context vectors:
\begin{equation}
c_{i}=\sum_{j=1}^{Tx}a_{ij}h_{i}^{m}
\end{equation}
\begin{equation}
a_{i,j}=\frac{h_{j}^{m}z_{i}^{m}}{\sum _{j}h_{j}^{m}z_{i}^{m}}
\end{equation}
$z_{i}^{k}$ is computed using the following formula:
\begin{equation}
z_{j}^{k}=LSTM(z_{j-1}^{k},z_{j}^{k-1})
\end{equation}

\section{Phrase Table as Recommendation Memory for NMT}

In section 2 we described how the standard NMT models calculate the probability of the next target word (Eq. (2)). Our goal in this paper is to improve the accuracy of this probability estimation by incorporating information from phrase tables. Our main idea is to find the recommendable words and increase their probabilities at each decoding time step. Thus, three questions arise:

1) Which words are worthy to recommend at each decoding step?

2) How to determine an appropriate bonus value for each recommendable word?

3) How to integrate the bonus value into NMT?

In this section, we will describe the specific methods to answer above three questions. As the basis of our work, we first introduce two definitions used by our methods. 

\textbf { Definition 1 (prefix of phrase):} the prefix of a phrase is a word sequence which begins with the first word of the phrase and ends with any word of the phrase. Note that the prefix string can be empty. For a phrase $E=(e_{1},e_{2},e_{3})$, this phrase contains four prefixes: $\left \{ \emptyset, e_{1}, e_{1}e_{2}, e_{1}e_{2}e_{3} \right \}$.

\textbf {Definition 2 (suffix of partial translation):} the suffix of the partial translation $y_{<i}$ is a word sequence, which begins with any word belonging to $y_{<i}$, and ends with $y_{i-1}$. Similarly, the suffix string can also be empty. For partial translation $y_{<4}=( y_{1},y_{2},y_{3} )$, there are four suffixes $\left \{ \emptyset , y_{3} ,  y_{2}y_{3},  y_{1}y_{2}y_{3}  \right \}$.

\subsection{Word Recommendation Set}
\subsubsection{Candidate Target Phrase Set}
\noindent The first step is to derive a candidate target phrase set for a source sentence. The recommendation words are selected from this set.

Given a source sentence $X$ and a phrase translation table (as shown in upper right of Fig. 2), we can traverse the phrase translation table and get all the phrase pairs whose source side matches the input source sentence. Then, for each phrase pair, we add the target phrases with the top $N$ highest phrase translation probabilities into the candidate target phrase set.

In order to improve efficiency of the next step, we represent this candidate target phrase set in a form of prefix tree. If the phrases contain the same prefix (\textbf{Definition 1}), the prefix tree can merge them and represent them using the same non-terminal nodes. The root of this prefix tree is an empty node. Fig. 2 shows an example to illustrate how we get the candidate target phrase set for a source sentence. In this example, In phrase table (upper right), we find four phrases whose source side matches the source sentence (upper left). We add the target phrases into candidate target phrase set (middle). Finally, we use a prefix tree (bottom) to represent the candidate target phrases.  

\begin{figure}[!t]
	\centering
	\includegraphics[width=1\columnwidth]{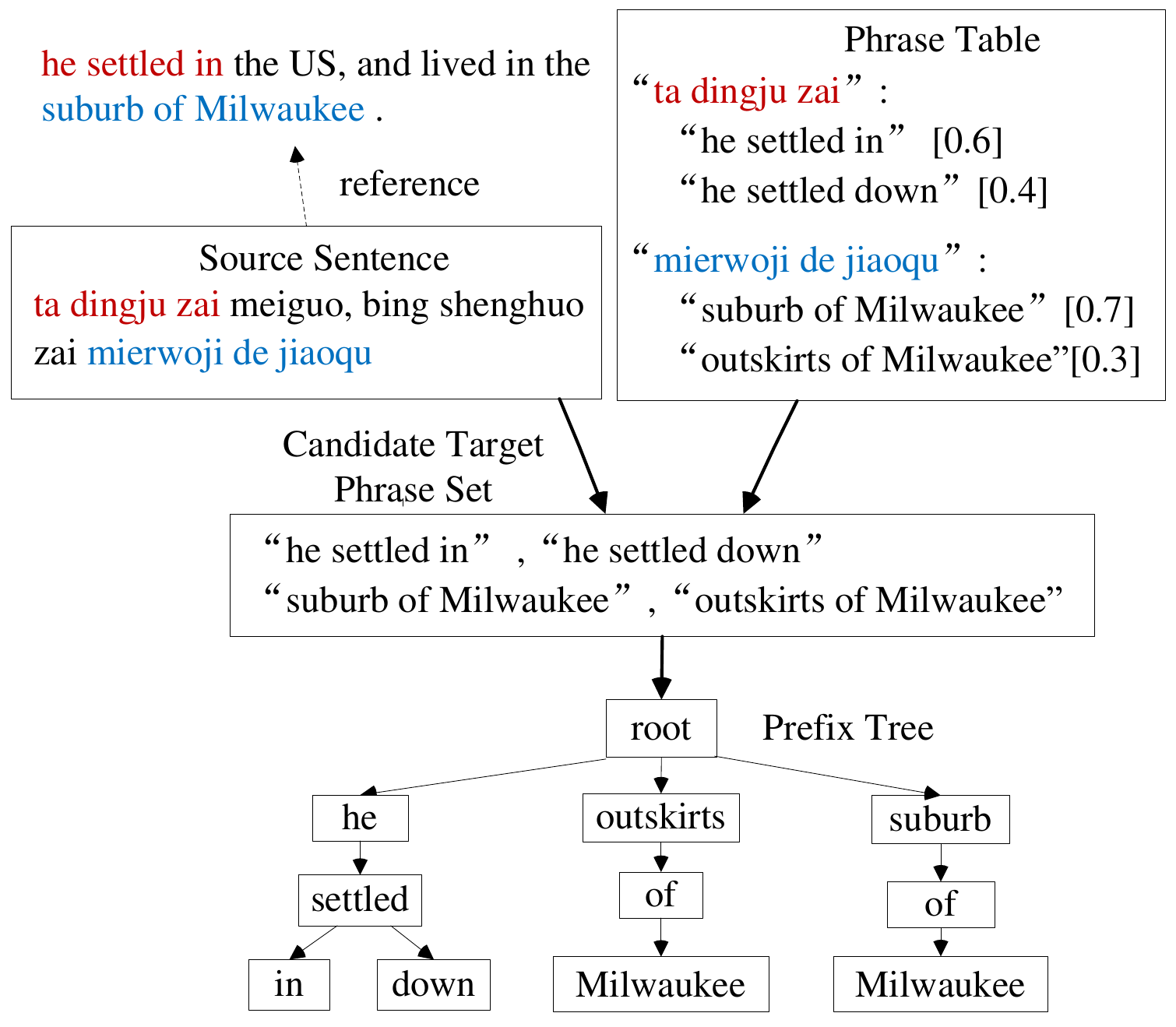}
	\caption{The procedure of constructing the target side prefix tree from candidate target phrase set for a source sentence.}
	\label{overview}
\end{figure}
\subsubsection{Word Recommendation Set}

\noindent With above preparations, we can start to construct the word recommendation set. In our method, we need to construct a word recommendation set $R_{i} $ at each decoding step $i$. The basic idea is that if a prefix $pf_{k}$ (\textbf{Definition 1}) of a phrase in candidate target phrase set matches a suffix $sf_{j}$ (\textbf{Definition 2}) of the partial translation $y_{<i}$, the next word of $pf_{k}$ in the phrase may be the next target word $y_{i}$ to be predicted and thus is worthy to recommend.

Here, we take Fig. 2 as an example to illustrate our idea. We assume that the partial translation is ``\emph{he settled in the US, and lived in the suburb of}''. According to our definition, this partial translation contains a suffix ``\emph{suburb of}''. Meanwhile, in candidate target phrase set, there is a phrase (``\emph{suburb of Milwaukee}'') whose two-word prefix is \emph{``suburb of''} as well. We can notice that the next word of the prefix ("Milwaukee") is exactly the one that should be predicted by the decoder. Thus, we recommend \emph{``Milwaukee''} by adding a bonus to it with the hope that when this low-frequency word is mistranslated by NMT, our recommendation can fix this mistake. 

Under this assumption, the procedure of constructing the word recommendation set $R_{i} $ is illustrated in \textbf{Algorithm 1}. We first get all suffixes of $y_{<i}$ (line 2) and all prefixes of target phrases belonging to candidate target phrase set (line 3). If a prefix of the candidate phrase matches a suffix of $y_{<i}$, we add the next word of the prefix in the phrase into recommendation set  $R_{i} $ (line 4-7). 

In the definition of the prefix and suffix, we also allow them to be an empty string. By doing so, we can add the first word of each phrase into the word recommendation set, since the suffix of $y_{<i}$ and the prefix of any target phrase always contain a match part $\emptyset$.
The reason we add the first word of the phrase into recommendation set is that we hope our methods can still recommend some possible words when NMT has finished the translation of one phrase and begins to translate another new one, or predicts the first target word of the whole sentence.

\begin{algorithm}[t]
	\caption{Construct recommendation word set}  
	\hspace*{0.03in} {\bf Input:} 
	candidate target phrase set;
	already generated partial translation $y_{<i}$\\
	\hspace*{0.03in} {\bf Output:}  
	word recommendation set $R_{i} $
	\begin{algorithmic}[1]
		\State $R_{i}=\left \{  \right \}$
		\State Get all suffixes of $y_{<i}$ (denote each suffix by $sf_{j}$)
		\State Get all prefixes of each target phrase in candidate target phrase set (denote every prefix by $pf_{k}$ )
		\For{each suffix $sf_{j}$ and each prefix $pf_{k}$}
		\If{ $sf_{j}==pf_{k}$} 
		\State Add the next word of $pf_{k}$ into $R_{i} $
		\EndIf
		\EndFor
		\State \Return  $ R_{i}$
	\end{algorithmic}
\end{algorithm}

Now we already know which word is worthy to recommend. In order to facilitate the calculation of the bonus value (section 3.2), we also need to maintain the origin of each recommendation word. Here, the origin of a recommendation word contains two parts: 1) the phrase pair this word belongs to and 2) the phrase translation probability between the source and target phrases. Formally, for a recommendation word $t$, we can denote it by:
\begin{equation}
t \in \left \{ (E_{t}, F_{t})^{m},\ p^{m}_{pht}(E_{t},F_{t}) \\  \right \}_{m=1}^{M}
\end{equation}
where $(E_{t},F_{t})^{m} $ denotes the $m$-th phrase pair the recommendation word $t$ belongs to (some words may belong to different phrase pairs and $M$ denotes the number of phrase pairs). $E_{t}$ is the source phrase and $F_{t}$ is the target phrase. $p^{m}_{pht}(E_{t},F_{t})$ is the phrase translation probability between the source and target phrases\footnote{Here the phrase translation probability is the mean of four probabilities, i.e., the bidirectional phrase translation probabilities and bidirectional lexical terms.}. Take Fig. 2 as an example. When the partial translation is ``\emph{he}'', word \emph{``settled''} can be recommended according to algorithm 1. Word \emph{``settled''} is contained in two phrase pairs and the translation probabilities are respectively 0.6 and 0.4. Thus, we can denote the word "settled" as follows:
\begin{equation}
\text{settled} \in \begin{Bmatrix}
(\text{ta \ dingju \ zai}, \text{he settled\ in}),\ 0.6\\ 
(\text{ta \ dingju \ zai}, \text{he settled\ down}),\ 0.4
\end{Bmatrix}
\end{equation}

\subsection{Bonus Value Calculation}
\label{sect:pdf}
The next task is to calculate the bonus value for each recommendation word. For a recommendation word $t$ denoted by Eq. (8), its bonus value is calculated as follows:

\textbf{Step1:} Extracting each phrase translation probability $p^{m}_{pht}(E_{t}, F_{t})$. 

\textbf{Step2:} For each phrase pair $(E_{t}, F_{t})^{m}$, we convert the attention weight $a_{i,j}$ in NMT (Eq. (6)) between target word $y_{i}$ and source word $x_{j}$ to phrase alignment probability $a^{m}_{(i,E_{t})}$ between target word $y_{i}$ and source phrase $E_{t}$ as follows:

\begin{equation}
a^{m}_{(i,E_{t})}=\frac{\sum _{x_{j}\in E_{t}}a_{i,j}}{|E_{t}|}
\end{equation}
where $|E_{t}|$ is the number of words in phrase $E_{t}$. As shown in Eq. (10), our conversion method is making an average of word alignment probability $a_{i,j}$ whose source word $x_{j}$ belongs to source phrase $E_{t}$.

\textbf{Step3:} Calculating the bonus value for each recommendation word as follows:
\begin{equation}
V(R_{i})=\sum_{m=1}^{M}a^{m}_{(i,E_{t})}p^{m}_{pht}(E_{t},F_{t}) \ \ \text{if} \ t\in R_{i}
\end{equation}
From Eq. (11), the bonus value is determined by two factors, i.e., 1) alignment information $a_{(i,E_{t})}$ and 2) translation probability $p_{pht}(E_{t}, F_{t})$. The process of involving $a_{(i,E_{t})}$ is important because the bonus value will be influenced by different source phrases that systems focus on. And we take $p_{pht}(E_{t}, F_{t})$ into consideration with a hope that the larger $p_{pht}(E_{t}, F_{t})$ is, the larger its bonus value is. 

\subsection{Integrating Bonus Values into NMT}
\label{ssec:layout}
The last step is to combine the bonus value with the conditional probability of the baseline NMT model (Eq.(2)). Specifically, we add the bonuses to the words on the basis of original NMT score (Eq. (3)) as follows:
\begin{equation}
\begin{aligned}
&p(y_{i}|c_{i},y_{<i})=p(y_{i}|c_{i},y_{<i},R_{i})\\&
=softmax((1+\lambda V(R_{i}))score(W_{y},\widetilde{z_{i}}))
\end{aligned}
\end{equation}
where $V(R_{i})$ is calculated by Eq. (11). $\lambda$ is the bonus weight, and specifically, it is the result of sigmoid function ($\lambda=sigmoid(x)$), where $x$ is a learnable parameter, and this sigmoid function ensures that the final weight falls between 0 and 1\footnote{In our preliminary experiments, we also try another strategy which adds the bonus to the NMT results as a bias, while the performance of this strategy is lower than the current introduced method (Eq. 12).}.

% Removed by KO  we are not accepting printed papers any more!!!
%  Exceptionally,
% authors for whom it is \emph{impossible} to print on A4 paper may use
% \emph{US Letter} paper. In this case, they should keep the \emph{top}
% and \emph{left} margins as given above, use the same column width,
% height and gap, and modify the bottom and right margins as
% necessary. Note that the text will no longer be centered.
\section{Experimental Settings}
In this section, we describe the experiments to evaluate our proposed methods. 
\subsection{Dataset}
We test the proposed methods on Chinese-to-English (CH-EN) translation and English-to-Japanese (EN-JA) translation. In CH-EN translation, we test the proposed methods with two data sets: 1) small data set, which includes 0.63M\footnote{LDC2000T50, LDC2002L27, LDC2002T01, LDC2002E18,  LDC2003E07, LDC2003E14, LDC2003T17, LDC2004T07.} sentence pairs; 2) large-scale data set, which contains about 2.1M sentence pairs. NIST 2003 (MT03) dataset is used for validation.  NIST2004-2006 (MT04-06) and NIST 2008 (MT08) datasets are used for testing. In EN-JA translation, we use KFTT dataset\footnote{http://www.phontron.com/kftt/.}, which includes 0.44M sentence pairs for training, 1166 sentence pairs for validation and 1160 sentence pairs for testing. 

\subsection{Training and Evaluation Details}
We use the Zoph\_RNN toolkit\footnote{\url{https://github.com/isi-nlp/Zoph RNN}. We extend this toolkit with global attention.} to implement all our described methods. In all experiments, the encoder and decoder include two stacked LSTM layers. The word embedding dimension and the size of hidden layers are both set to 1,000. The minibatch size is set to 128. We limit the vocabulary to 30K most frequent words for both the source and target languages. Other words are replaced by a special symbol ``UNK''. At test time, we employ beam search and beam size is set to 12. We use case-insensitive 4-gram BLEU score as the automatic metric \cite{papineni2002bleu} for translation quality evaluation.

\subsection{Phrase Translation Table}
Our phrase translation table is learned directly from parallel data by Moses \cite{koehn2007moses}. To ensure the quality of the phrase pair, in all experiments, the phrase translation table is filtered as follows: 1) out-of-vocabulary words in the phrase table are replaced by UNK; 2) we remove the phrase pairs whose words are all punctuations and UNK; 3) for a source phrase, we retain at most 10 target phrases having the highest phrase translation probabilities.

\subsection{Translation Methods}

\begin{table*}[htbp]
\centering
 \begin{tabular}{|c|c|c|cccc|c|c|c|}
  \hline
  \multirow{2}{*}{\#} & \multirow{2}{*}{\textbf{Method}} &  \multicolumn{6}{|c|}{\textbf{CH-EN}} & \multicolumn{2}{|c|}{\textbf{EN-JA}} \\
  \cline{3-10}
  &  & \textbf{MT03(dev)} & \textbf{MT04} & \textbf{MT05} & \textbf{MT06} &\textbf{MT08} & \textbf{Ave} & \textbf{dev} & \textbf{test}\\
  \hline
1 & \textbf{Moses}             & 28.35 & 30.02 & 29.10 & 32.92 & 23.20 & 28.72 &20.06&22.40\\
2 & \textbf{Baseline}          & 34.20 & 36.96 & 32.60 & 33.85 & 25.96 & 32.71
&23.61 &25.99\\

3 & \textbf{Arthur} & 34.98$^{\dagger}$ & 37.96$^{\dagger}$ & 33.36$^{\dagger}$ & 34.79$^{\dagger}$ & 26.53$^{*}$& 33.52 & 24.33$^{*}$& 26.72$^{\dagger}$\\
  \hline

4 & \textbf{Our method}     & \textbf{36.48}$^{\dagger}$ & \textbf{38.79}$^{\dagger}$ & \textbf{35.34}$^{\dagger}$ & \textbf{36.58}$^{\dagger}$ & \textbf{27.49}$^{\dagger}$ & \textbf{34.94}
&\textbf{25.63}$^{\dagger}$& \textbf{27.95}$^{\dagger}$\\
  \hline

5 & \textbf{system(no matching)}     & 34.99$^{\dagger}$ & 37.54$^{*}$ & 33.32$^{\dagger}$ & 34.22$^{*}$ & 26.39$^{*}$ 
& 33.29 &24.11$^{*}$& 26.47$^{*}$\\

6 & \textbf{system(no first)}    & 35.25$^{\dagger}$ & 38.07$^{\dagger}$ & 34.13$^{\dagger}$ & 34.95$^{\dagger}$ & 26.67$^{\dagger}$ & 33.81 &24.37$^{\dagger}$& 26.93$^{\dagger}$\\

  \bottomrule
 \end{tabular}
\caption{\label{tab:test} Translation results (BLEU score) for different translation methods. ``$*$'' indicates that it is statistically significant better ($p<0.05$) than Baseline and ``${\dagger}$'' indicates $p<0.01$.}
\end{table*} 

We compare our method with other relevant methods as follows:

1) \textbf{Moses}: It is a widely used phrasal SMT system \cite{koehn2007moses}.

2) \textbf{Baseline}: It is the baseline attention-based NMT system \cite{luongeffective,zoph2016multi}.

3) \textbf{Arthur}: It is the state-of-the-art method which incorporates discrete translation lexicons into NMT model \cite{arthur2016incorporating}. We choose automatically learned lexicons and bias method. We implement the method on the base of the baseline attention-based NMT system. Hyper parameter $\varepsilon$ is 0.001, the same as that reported in their work.

\section{Translation Results}
Table 1 reports the detailed translation results for different methods. Comparing the first two rows in Table 1, it is very obvious that the attention-based NMT system \textbf{Baseline} substantially outperforms the phrase-based SMT system \textbf{Moses} on both CH-EN translation and EN-JA translation. The average improvement for CH-EN and EN-JA translation is up to 3.99 BLEU points (32.71 vs. 28.72) and 3.59 BLEU (25.99 vs. 22.40) points, respectively.

\subsection{Effect of Integrating Phrase Translation Table}
The first question we are interested in is whether or not phrase translation table can improve the translation quality of NMT. Compared to the baseline, our method markedly improves the translation quality on both CH-EN translation and EN-JA translation. In CH-EN translation, the average improvement is up to 2.23 BLEU points (34.94 vs. 32.71). In EN-JA translation, the improvement can reach 1.96 BLEU points (27.95  vs. 25.99). It indicates that incorporating a phrase table into NMT can substantially improve NMT's translation quality.

\begin{figure}[!t]
	\centering
	\includegraphics[width=0.9\columnwidth]{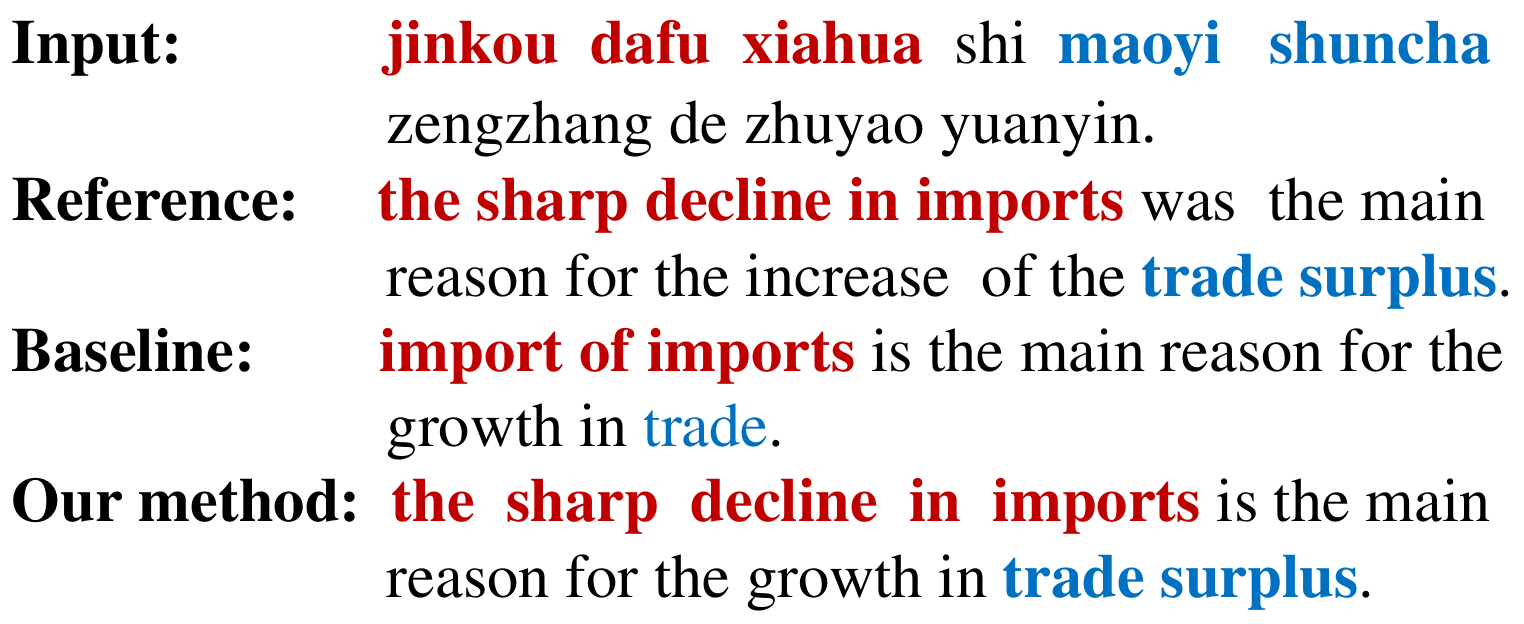}
	\caption{Translation examples, where the proposed method is able to obtain a correct translation while the baseline NMT is not. }
	\label{overview}
\end{figure}

In Fig. 3, we show an illustrative example of CH-EN translation. In this example, our method is able to obtain a correct translation while the baseline is not. Specifically, baseline NMT system mistranslates ``jinkou dafu xiahua (the sharp decline in imports)'' into ``import of imports'', and incorrectly translates ``maoyi shuncha (trade surplus)'' into ``trade''. But these two mistakes are fixed by our method, because there are two phrase translation pairs (``jinkou dafu xiahua'' to ``the sharp decline in imports'' and ``maoyi shuncha'' to ``trade surplus'') in the phrase table, and the correct translations are obtained due to our recommendation method.

\subsection{Lexicon vs. Phrase}

A natural question arises that whether it is more beneficial to incorporate a phrase translation table than the translation lexicons. From Table 1, we can conclude that both translation lexicons and phrase translation table can improve NMT system's translation quality. In CH-EN translation, \textbf{Arthur} improves the baseline NMT system with 0.81 BLEU points, while our method improves the baseline NMT system with 2.23 BLEU points. In EN-JA translation, \textbf{Arthur} improves the baseline NMT system with 0.73 BLEU points, while our method improves the baseline NMT system with 1.96 BLEU points. Therefore, it is very obvious that phrase information is more effective than lexicon information when we use them to improve the NMT system.

\begin{table}[htbp]
	\centering
	\begin{tabular}{cc}
		\toprule
		\textbf{Method} & \textbf{Faithfulness }  \\
		\midrule
		\textbf{Baseline} &  3.21 \\
		\textbf{Arthur}  & 3.25  \\
		\textbf{Our method}  &  3.33 \\
		\bottomrule
	\end{tabular}
	\caption{\label{tab:test} Subjective evaluation of translation faithfulness.}
\end{table}

Fig. 4 shows an illustrative example. In this example, baseline NMT mistranslates ``dianli (electricity) anquan (safe)'' into ``coal''. \textbf{Arthur} partially fixes this error and it can correctly translate ``dianli (electrical)'' into ``electrical'', but the source word ``anquan (safe)'' is still missed. Fortunately, this mistake is fixed by our proposed method. The reason behind this is that \textbf{Arthur} uses information from translation lexicons, which makes the system only fix the translation mistake of an individual lexicon (in this example, it is ``dianli (electrical)''), while our method uses the information from phrases, which makes the system can not only obtain the correct translation of the individual lexicon but also capture local lexicon reordering and fixed collocation etc.

Besides	the BLEU score, we also conduct a subjective evaluation to validate the benefit of incorporating a phrase table in NMT. The subjective evaluation is conducted on CH-EN translation. As our method tries to solve the problem that NMT system cannot reflect the true meaning of the source sentence, the criterion of the subjective evaluation is the faithfulness of translation results. Specifically, five human evaluators, who are native Chinese and expert in English, are asked to evaluate the translations of 500 source sentences randomly sampled from the test sets without knowing which system a translation is selected from. The score ranges from 0 to 5. For a translation result, the higher its score is, the more faithful it is. Table 2 shows the average results of five subjective evaluations on CH-EN translation. As shown in Table 2，the faithfulness of translation results produced by our method is better than \textbf{Arthur} and baseline NMT system.

\begin{table*}[htbp]
	\centering
	\begin{tabular}{ccccccc}
		\toprule
		\textbf{Method} & \textbf{MT03} & \textbf{MT04} & \textbf{MT05} & \textbf{MT06} & \textbf{MT08} & \textbf{Ave} \\
		\midrule
		
		\textbf{Baseline}      & 39.07 & 40.49 & 37.26 & 38.04 & 28.83 & 36.74\\
		\textbf{Arthur} & 39.92$^{\dagger}$ & 41.41$^{\dagger}$ & 38.18$^{\dagger}$ & 38.67$^{\dagger}$ & 29.32$^{\dagger}$ & 37.50\\
		\textbf{Our method} & 40.87$^{\dagger}$ & 42.41$^{\dagger}$ & 39.29$^{\dagger}$ & 39.83$^{\dagger}$ & 30.47$^{\dagger}$ & 38.57\\
		
		\bottomrule
	\end{tabular}
	\caption{\label{tab:test} Translation results (BLEU score) for different translation methods on large-scale data. ``${\dagger}$'' indicates that it is statistically significant better ($p<0.01$) than Baseline.}
\end{table*}

\begin{figure}[!t]
	\centering
	\includegraphics[width=0.9\columnwidth]{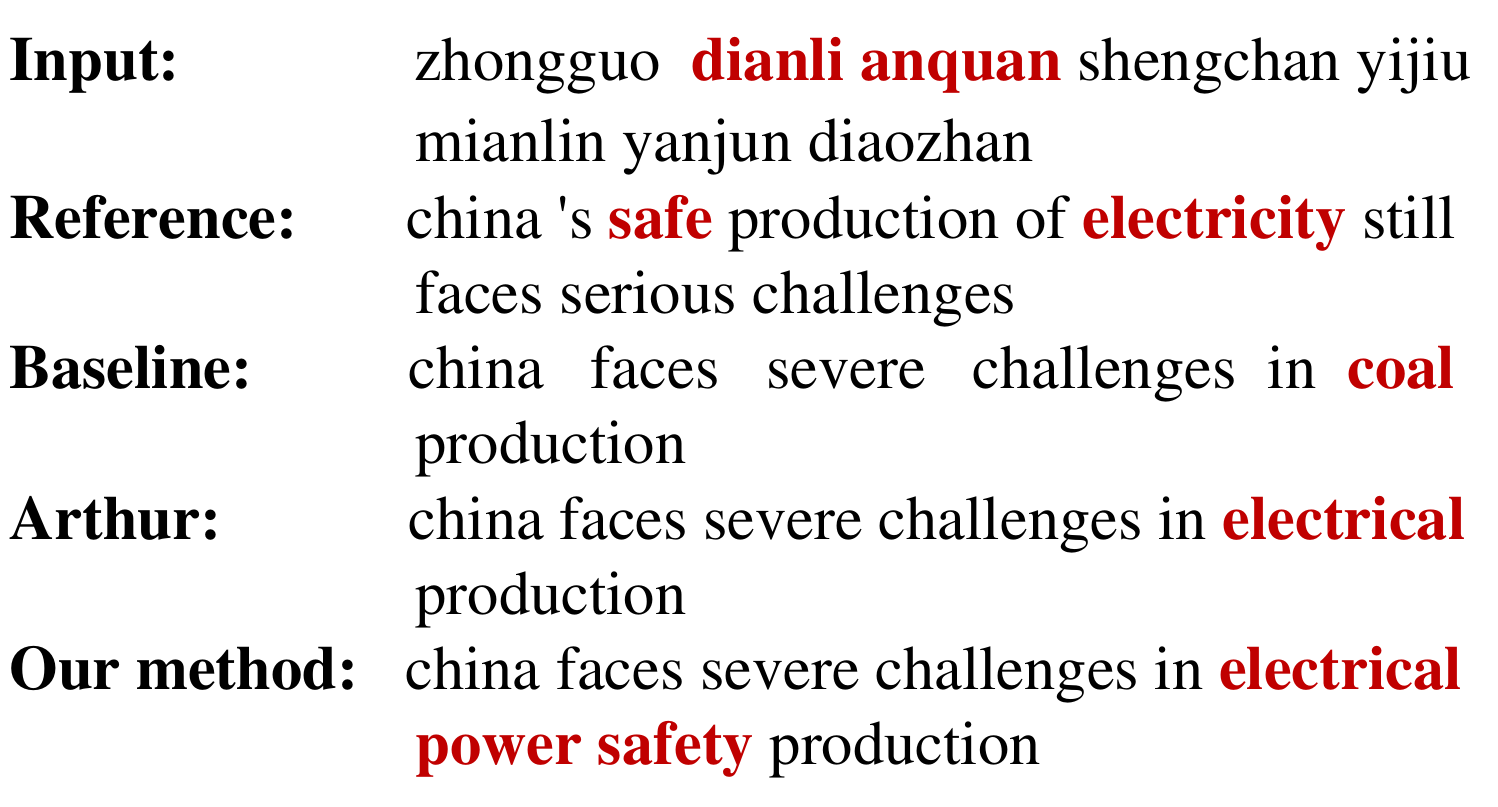}
	\caption{Translation examples, where both two methods can improve the baseline system, but our proposed model produces a better translation result. }
	\label{overview}
\end{figure}

\subsection{Different Methods to Construct Recommendation Set}

When constructing the word recommendation set, our current methods are adding the next word of the match part into recommendation set. In order to test the validity of this strategy, we compare the current strategy with another system, in which, we can add all words in candidate target phrase set into recommendation set without matching. We denote this system by \textbf{system(no matching)}, whose results are reported in line 5 in Table 1. From the results, we can conclude that in both CH-EN translation and EN-JA translation, \textbf{system(no matching)} can boost the baseline system, while the improvements are much smaller than our methods. It indicates that the matching between the phrase and partial translation is quite necessary for our methods.

As we discussed in Section 3.1, we allow the prefix and suffix to be an empty string to make first word of each phrase into the word recommendation set. To show effectiveness of this setting, we also implement another system as a comparison. In the system, the first words of each phrase are not included in the recommendation set (we denote the system by \textbf{system(no first)}). The results of this system are reported in line 6 in Table 1. As shown in Table 1,  our methods performs better than \textbf{system(no first)}) on both CH-EN translation and EN-JA translation. This result shows that the first word of the target phrase is also important for our method and is worthy to recommend.

\subsection{Translation Results on Large Data}

We also conduct another experiment to find out whether or not  our methods are still effective when much more sentence pairs are available. Therefore, the CH-EN experiments on millions of sentence pairs are conducted and Table 3 reports the results. We can conclude from Table 3 that our model can also improve the NMT translation quality on all of the test sets and the average improvement is up to 1.83 BLEU points.

\section{Related Work}
\label{sec:length}

In this work, we focus on integrating the phrase translation table of SMT into NMT. And there have been several effective works to combine SMT and NMT.

\textbf{Using coverage mechanism.} \cite{tu2016coverage} and \cite{mi2016coverage} improved the over-translation and under-translation problems in NMT inspired by the coverage mechanism in SMT.

\textbf{Extending beam search.} \cite{Dahlmann2017Hyibrid} extended the beam search method with SMT hypotheses. \cite{stahlberg2016syntactically} improved the beam search by using the SMT lattices. 

\textbf{Combining SMT features and results}.
\cite{he2016improved} presented a log-linear model to integrate SMT features (translation model and the
language model) into NMT.
\cite{liu2016supervised} and \cite{mi2016supervised} proposed a supervised attention model for NMT to minimize the alignment disagreement between NMT and SMT.
\cite{wang2016neural} proposed a method that incorporates the translations of SMT into NMT with an auxiliary classifier and a gating function. 
\cite{zhou2017combination} proposed a neural combination model to fuse the NMT translation results and SMT translation results. 

\textbf{Incorporating translation lexicons}.
\cite{arthur2016incorporating,feng2017memory} attempted to integrate NMT with the probabilistic translation lexicons. 
\cite{zhang2016bridging} moved forward further by incorporating a bilingual dictionaries in NMT.

In above works, integrating the phrase translation table of SMT into NMT has not been fully studied.

\textbf{Translating phrase in NMT}.
The most related works are \cite{tang2016neural} and \cite{Wang2017phrases}. Both methods attempted to explore the possibility of translating phrases as a whole in NMT. In their models, NMT can generate a target phrase in phrase memory or a word in vocabulary by using a gate. However, their ``phrases'' are different from that are used in phrase-based SMT. \cite{tang2016neural}'s models only support a unique translation for a source phrase. In \cite{Wang2017phrases}'s models, the source side of a phrase pair must be a chunk. Different from above two methods, our model can use any phrase pair in the phrase translation table and promising results can be achieved.

\section{Conclusions and Future Work}

In this paper, we have proposed a method to incorporate a phrase translation table as recommendation memory into NMT systems to alleviate the problem that the NMT system is opt to generate fluent but unfaithful translations.

Given a source sentence and a phrase translation table, we first construct a word recommendation set at each decoding step by using a matching method. Then we calculate a bonus value for each recommendable word. Finally we integrate the bonus value into NMT. The extensive experiments show that our method achieved substantial increases in both Chinese-English and English-Japanese translation tasks.

In the future, we plan to design more effective methods to calculate accurate bonus values.

\section*{Acknowledgments}
The research work described in this paper has been supported by the National Key Research and Development Program of China under Grant No. 2016QY02D0303 and the Natural Science Foundation of China under Grant No. 61333018 and 61673380. The research work in this paper also has been supported by Beijing Advanced Innovation Center for Language Resources. 

\bibliographystyle{named}
\bibliography{ijcai18}
\end{CJK*}
\end{document}